\documentclass[graybox]{svmult}

\usepackage[misc]{ifsym}
\usepackage{pgfplots}
\pgfplotsset{width=7cm,compat=1.8}
\usepackage{type1cm}        
\usepackage{makeidx}         
\usepackage{graphicx}        
\usepackage{multicol}        
\usepackage[bottom]{footmisc}

\usepackage{newtxtext}       %
\usepackage[varvw]{newtxmath}       
\usepackage[T1]{fontenc}
\usepackage{listings}
\usepackage{booktabs}
\usepackage{caption}

\usepackage{amsmath}
\usepackage[caption=false]{subfig}

\makeindex             

\begin{document}

\title*{Towards Trust of Explainable AI in Thyroid Nodule Diagnosis}
\author{Truong Thanh Hung Nguyen, Van Binh Truong, Vo Thanh Khang Nguyen, Quoc Hung Cao, Quoc Khanh Nguyen}
\authorrunning{T. T. H. Nguyen, V. B. Truong, V. T. K. Nguyen, Q. H. Cao, Q. K. Nguyen}
\institute{T. T. H. Nguyen (\Letter) \at Friedrich-Alexander-Universität Erlangen-Nürnberg, Erlangen, Germany\\ \email{hung.tt.nguyen@fau.de}
\and 
V. B. Truong \and V. T. K. Nguyen \and Q. H. Cao \and Q. K. Nguyen \and T. T. H. Nguyen \at Quy Nhon AI, FPT Software, Quy Nhon, Vietnam \\ \email{binhtv8@fsoft.com.vn} 
\and 
V. T. K. Nguyen \at \email{khangnvt1@fsoft.com.vn}
\and
Q. H. Cao \at \email{hungcq3@fsoft.com.vn} 
\and 
Q. K. Nguyen \at \email{khanhnq33@fsoft.com.vn}
}
%
%
\maketitle

\abstract*{The ability to explain the prediction of deep learning models to end-users is an important feature to leverage the power of artificial intelligence (AI) for the medical decision-making process, which is usually considered non-transparent and challenging to comprehend.
In this paper, we apply state-of-the-art eXplainable artificial intelligence (XAI) methods to explain the prediction of the black-box AI models in the thyroid nodule diagnosis application. 
We propose new statistic-based XAI methods, namely Kernel Density Estimation and Density map, to explain the case of no nodule detected.
XAI methods' performances are considered under a qualitative and quantitative comparison as feedback to improve the data quality and the model performance.
Finally, we survey to assess doctors' and patients' trust in XAI explanations of the model's decisions on thyroid nodule images.}

\abstract{The ability to explain the prediction of deep learning models to end-users is an important feature to leverage the power of artificial intelligence (AI) for the medical decision-making process, which is usually considered non-transparent and challenging to comprehend.
In this paper, we apply state-of-the-art eXplainable artificial intelligence (XAI) methods to explain the prediction of the black-box AI models in the thyroid nodule diagnosis application. 
We propose new statistic-based XAI methods, namely Kernel Density Estimation and Density map, to explain the case of no nodule detected.
XAI methods' performances are considered under a qualitative and quantitative comparison as feedback to improve the data quality and the model performance.
Finally, we survey to assess doctors' and patients' trust in XAI explanations of the model's decisions on thyroid nodule images.}
\section{Introduction}
\label{sec:introduction}
Thyroid cancer is one of the most common cancer types and is the leading cause of cancer death worldwide~\cite{deng2020global, kim2020geographic}, especially during the COVID-19 pandemic~\cite{giannoula2022updated, guven2021prognostic}. 
Characterized by malignant cells formed in the thyroid gland tissues, the thyroid cancer prognosis depends on the type and the stage at which the disease is detected.
Often, doctors rely on the medical images' interpretation, such as thyroid ultrasound images, to identify nodules' presence and provide a diagnosis. 
However, in routine cancer screening, the errors are mainly false negatives, in which a nodule is present but undetected~\cite{hebert_redundancy_2020}.
Due to recent advances in AI, deep learning models can now serve as decision support means for medical experts.
A medical diagnosis system must be accurate, transparent, and explainable to gain end-users' trust.
Considering the explainability capability, simple AI methods such as linear regression and decision trees are self-explanatory. 
Still, these methods lack the complexity required for tasks such as classifying two and three-dimensional medical images.
Given the increasing ubiquity of advanced techniques such as deep neural networks (DNNs), a new challenge for medical AI is its so-called black-box nature, with decisions that seem opaque and inscrutable, even for experts to understand~\cite{10.1145/3236009}.
Thus, while their opacity is deeply intertwined with their success, it poses a challenge for applying DNNs to high-stakes problems such as medical imaging until we can develop methods that allow radiologists to develop understanding and appropriate trust.
Furthermore, newer regulations like the European General Data Protection Regulation (GDPR) strictly require transparency in black-box models, especially in healthcare. 
Thus, there is a growing chorus of researchers calling for XAI methods. 
Therefore, in this paper, our main contributions are:
\begin{enumerate}
    \item We applied several XAI methods, namely LIME~\cite{Lime}, RISE~\cite{petsiuk2018rise}, Grad-CAM~\cite{selvaraju2017grad}, Grad-CAM++~\cite{gradcamplusplus}, Ada-SISE~\cite{sudhakar2021ada}, LRP~\cite{bach2015pixel}, and D-RISE~\cite{petsiuk2021black} to explain the two-stage model's classification and localization of nodules.
    \item We proposed two statistic-based XAI methods, namely Kernel Density Estimation (KDE) and Density map (DM), to monitor the two-stage model's localization process from the first stage to the second stage and further explain the case of no nodule detected, especially false negative case.
    \item We evaluated XAI methods' performance and suitability for the specific nodule detection cases with qualitative and quantitative results and a surveyed XAI's trust to end-users.
\end{enumerate}
Our code is available for reproductivity on GitHub\footnote{\url{https://github.com/hungntt/xai\_thyroid}}.

\subsection{Dataset}\label{subsec:dataset}
\begin{figure}[b]
\sidecaption
    \subfloat[With nodule]{\includegraphics[width=0.28\hsize]{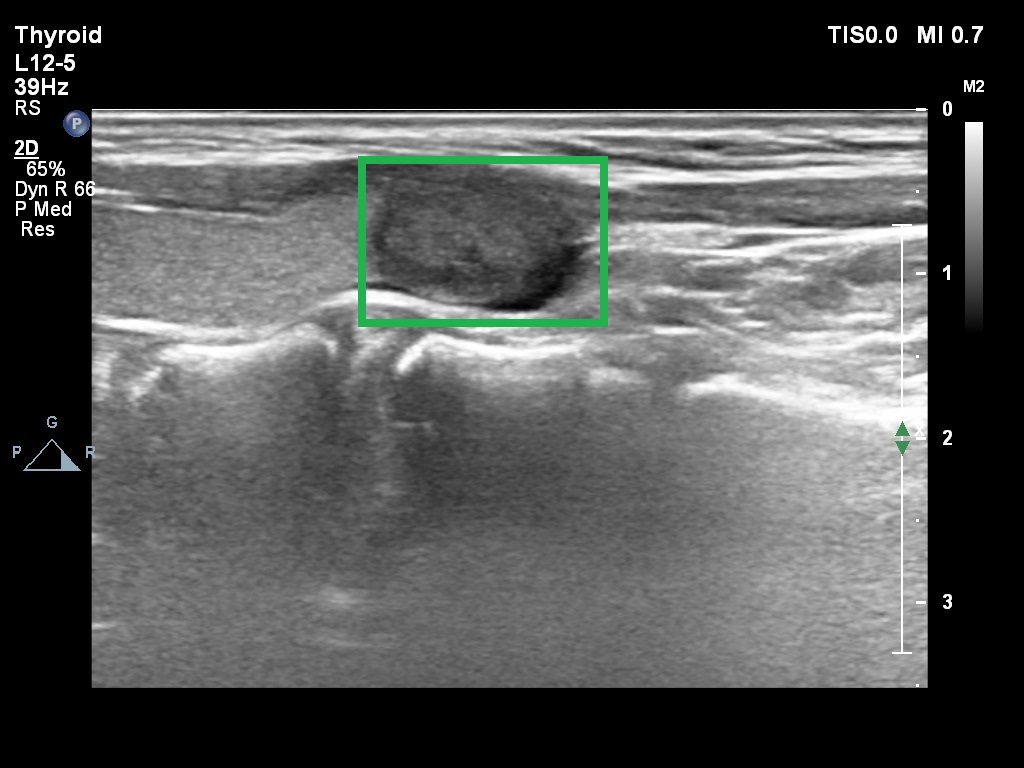}}
    \hfill
    \subfloat[Without nodule]{\includegraphics[width=0.28\hsize]{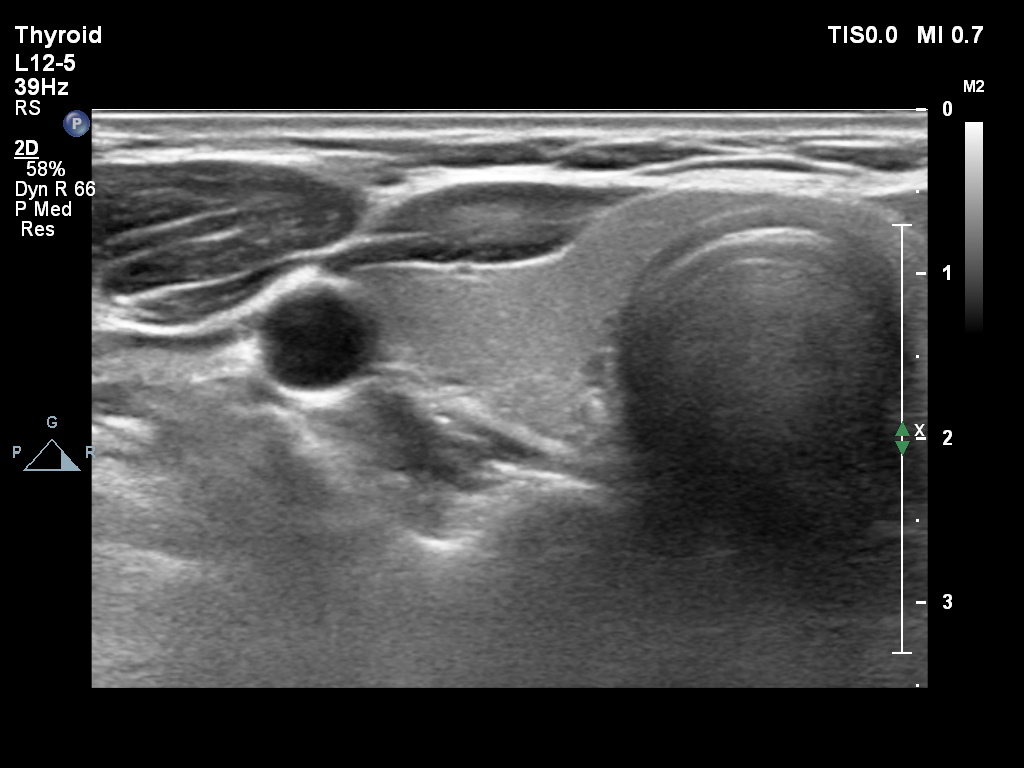}}
\caption{Examples of the Vietnamese thyroid ultrasound dataset. The green box is the ground truth labeled by doctors.}
\label{fig:sample_image}  
\end{figure}

We use the thyroid ultrasound dataset from~\cite{pham2021evaluating} that contains 14171 thyroid ultrasound images of 970 Vietnamese patients. Samples are shown in Fig.~\ref{fig:sample_image}, and medical experts label the nodule locations.

\section{Related Work}
\label{sec:related_work}
\subsection{Backpropagation-based methods}\label{subsec:backpropagation-based-methods}
Backpropagation-based methods calculate the gradients of the model's output to the input features or hidden neurons.
Hence, they utilize the backward pass of information flow to understand neuronal influence and relevance of input towards the output.
The first gradients explanation technique proposed in~\cite{DBLP:journals/corr/SimonyanVZ13} computes how much a change in each input dimension changes the predictions in a small neighborhood around the input.
Some preceding backpropagation-based equations take relative importance given to gradient value during backpropagation to generate saliency maps~\cite{10.1007/978-3-319-10590-1_53, smilkov2017smoothgrad}.
While LRP~\cite{bach2015pixel} modifies backpropagation rules to measure the relevance or irrelevance of the input features to the models' prediction.

\subsection{CAM-based methods}\label{subsec:cam-based-methods}
Based on the CAM method~\cite{zhou2016learning}, an extensive research effort on CAM-based methods has been put to blend high-level features extracted by CNNs into a unique explanation map.
Grad-CAM~\cite{selvaraju2017grad} and Grad-CAM++~\cite{gradcamplusplus} are two improvements over CAM that utilize backpropagation to provide a better visual explanation for classifiers.
 
\subsection{Perturbation-based methods}\label{subsec:perturbation-based-methods}
Perturbation-based methods are a class of techniques for explaining the decision-making process of DNNs by modifying the model's input and observing the output's changes. 
LIME~\cite{Lime} explains the prediction by learning an interpretable model that approximates the model locally around a data point using occlusions of superpixels. 
RISE~\cite{petsiuk2018rise} proposed a method for producing saliency maps using random perturbation techniques without having to analyze the model's complex structure. 
D-RISE~\cite{petsiuk2021black} extended RISE to produce saliency maps for object detectors.
SISE~\cite{sattarzadeh2021explaining} improved upon RISE's fidelity and plausibility using attribution-based input sampling techniques. 
Still, it has high computational costs when there are many activation maps with positive slopes that are inefficient in the prediction process. 
Ada-SISE~\cite{sudhakar2021ada} was developed to solve this problem by removing unnecessary objects, which saves computational time and provides a better reasonable explanation.

\subsection{Statistic-based methods}\label{subsec:statistic-based-methods}
Kernel Density Estimation (KDE) is a non-parametric mathematical method for estimating the probability density function of a continuous variable~\cite{terrell1992variable, wkeglarczyk2018kernel} which is one of the most common methods for estimating density level, set estimation, clustering, or unsupervised learning~\cite{parzen1962estimation}.
Recently, KDE has been made explainable with LRP for outlier and inlier detection in unsupervised learning models without the ground-truth labels~\cite{montavon2022explaining}.
Density map is commonly used in crowd counting literature, which is usually for estimating the distribution of objects, namely people in the scene~\cite{li2020density}.
However, the idea of counting the model's detected boxes to estimate the distribution of predicted boxes as an explanation has not been applied in previous works. 

\subsection{XAI in the medical diagnosis system}\label{subsec:xai-in-the-medical-diagnosis-system}
Several XAI applications for different cancer diagnoses are proposed to answer the black-box AIs.
In recent years, there have been 37 publications on how XAI is applied in skin cancer detection~\cite{skincancer}. More than half of the articles applied current XAI methods to their model, nearly 40\% tried to solve specific problems such as bias detection and the effect of XAI on man-machine interactions, and the remaining 10\% offered novel XAI methods or enhanced existing techniques.
Recently, during the outbreak of COVID-19, LIME is also applied to explain the model's interpretability for screening patients with COVID-19 symptoms~\cite{ahsan2020study}.
In the same context as our study, AIBx~\cite{thomas2020aibx} employed the image similarity model and physicians to create an XAI model, increasing physicians’ confidence in the predictions during the thyroid cancer diagnosis process.

However, the application of a wide-ranged number of XAI methods to nodule detection on thyroid ultrasound datasets has not been discovered yet.
Urgently, very little is known about the influence of XAI on the predictive performance, confidence, and model trust of doctors and radiologists in an artificial setting, and nothing is known about its effects in a clinical setting. 

\section{Methodology}
\label{sec:methodology}
\subsection{Object Detector and XAI categorization}
\subsubsection{Analysis of images with nodule}
\begin{figure}[bth!]
\sidecaption
\includegraphics[width=\hsize]{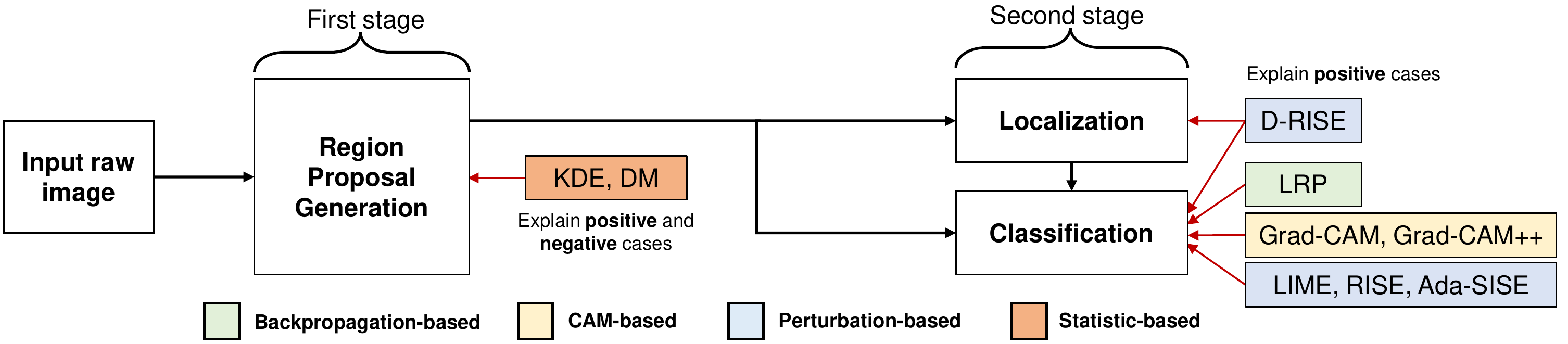}
\caption{Object detector's architecture and XAI methods' applicability to different tasks (red arrows).}
\label{fig:xai_categories}
\end{figure}
Our object detector employs state-of-the-art object detection networks based on the Faster R-CNN~\cite{fasterrcnn} and RetinaNet~\cite{RetinaNet} architectures to detect thyroid in ultrasound images.
The model detection process comprises two stages. 
In the first stage, the Region Proposal Network generates object proposals from input images. 
Next, a bounding box is predicted for each object proposal, with a probability of whether the box contains a thyroid nodule. 
In the second stage, the Region-of-Interest pooling layer implements bounding box regression and bounding box classifier. 
We categorize XAI methods in terms of their applicability to three main blocks of an object detector (as shown in Fig.~\ref{fig:xai_categories}):
\begin{itemize}
    \item Region Proposal Generation (Which proposals are generated by the model during the model's first stage?): Kernel Density Estimation (KDE), Density map (DM).
    \item Classification (Which features of an image make the model classify an image containing a nodule(s) at the model's second stage?): LRP, Grad-CAM, Grad-CAM++, LIME, RISE, Ada-SISE, D-RISE.
    \item Localization (Which features of an image does the model consider to detect a specific box containing a nodule at the model's second stage?): D-RISE.
\end{itemize}

Because XAI methods for the second stage (LRP, Grad-CAM, Grad-CAM++, LIME, RISE, Ada-SISE, D-RISE) require the model's output bounding boxes, they are only applicable to positive cases, where the model detects a nodule in the image.
While XAI methods for the first stage (KDE, DM) directly extract the model's attempts to find a nodule, they can be further applied to negative cases where the model does not detect any nodule.




\subsubsection{Local Interpretable Model-agnostic Explanation (LIME)}\label{subsubsec:local-interpretable-model-agnostic-explanation-(lime)}
For any given instance and its corresponding prediction, LIME utilizes Simple Linear Iterative Clustering~\cite{achanta2010slic} as a segmentation algorithm to randomly sample data around the neighborhood of the input instance for which produced predictions.
These generated data are used to train a local model.
The local model's prediction generates an explanation by weighting each sample according to the instance.
Then, LIME uses LASSO as a feature selection technique to choose the most important segments that contribute the most to the prediction for the explanation.

\subsubsection{Random Input Sampling for Explanation (RISE)}\label{subsubsec:random-input-sampling-for-explanation-(rise)}
In the masking generation, we firstly use binary-bit masking $(0,1)$ to generate $N=500$ samples, where masks $M=\left\{M_{i}, 1 \leq i \leq N\right\}$ for each superpixel with a probability $p=0.5$~\cite{petsiuk2018rise}. 
We set each sample's size as $8\times 8$.
We upsample these masks using bilinear interpolation to ensure all values are in the range $[0, 1]$.
Then, we feed samples into the model to get the bounding boxes and the corresponding score for each box $S^b$.
Finally, RISE sums up all the masks using the box scores, which are predicted on each sample as the weight of each mask to explain the target box from the input image in the form of a saliency map.

\subsubsection{Adaptive-Semantic Input Sampling for Explanation (Ada-SISE)}
Ada-SISE selects multiple layers of the model and extracts feature maps by feeding the input image into the model. 
Then, it samples subsets of the feature maps that contain the most important features by partially backpropagating the signal to the layer and calculating the average gradient scores for each feature map. 
It then collects all feature maps with positive scores and applies an Otsu-based threshold~\cite{otsu1979threshold} to remove those with lower scores. 
It produces attribution masks by bilinearly interpolating and normalizing the positive feature maps. 
For each selected layer, it obtains a layer visualization map by computing the weighted sum of the attribution masks. 
Finally, it combines obtained saliency maps in a fusion module to produce a final explanation.

\subsubsection{Gradient-weighted class activation mapping (Grad-CAM) and Grad-CAM++}\label{subsubsec:Grad-CAM-and-Grad-CAM++}
Grad-CAM~\cite{selvaraju2017grad} and Grad-CAM++~\cite{gradcamplusplus} are methods for producing a saliency map, which is a visual representation of the regions in an input image that is most important for a particular task or model.

Grad-CAM is a technique that uses the gradients of the target class with respect to the final convolutional layer of a CNN to produce a coarse localization map of the important regions in the input image. The map is then upsampled and weighted by the gradients to produce the final saliency map.

Grad-CAM++ is an extension of Grad-CAM that produces a more fine-grained and accurate localization map by using the gradients of the target class to the lower convolutional layers and the final convolutional layer. It also introduces a new way of weighting the upsampled map using a weighted combination of the gradients of the target class and the activations of the lower convolutional layers. The final saliency map is also produced by upsampling and weighting the localization map using the weighted combination of gradients and activations.

\subsubsection{Layer-wise Relevance Propagation (LRP)}\label{subsubsec:lrp}
LRP uses the weights and activations of a neural network to propagate relevance scores from the output layer back to the input layer according to the conservation rule, which states that a neuron that receives what is from the upper layer must be redistributed entirely to the lower layer in equal numbers. 
The relevance scores are propagated by calculating the quantity representing how much neuron $j$ contributes to making neuron $k$, where $j$ and $k$ are neurons in two consecutive layers of the neural network. The propagation procedure terminates once the input features have been reached. 
LRP has several propagation rules with different non-linear rectifiers. 
Still, we use the epsilon rule, which adds a small positive term in the denominator of the equation used to propagate the relevance scores, because it helps to solve cases where the denominator is zero and typically leads to sparser explanations in terms of input features and less noise.

\subsection{XAI methods for the localization task}\label{subsec:xai-methods-for-localization-problem}
\subsubsection{Kernel Density Estimation (KDE)}\label{subsubsec:kernel-density-estimation-(kde)}
KDE creates a distribution density map by weighting distances of all data points for each specific location along the distribution. 
If more data points are grouped locally, the estimation is higher as the probability of seeing a point at that location increases.

In the object detector's first stage, we apply KDE to the center points of 300 independent and identically distributed $(b_0, b_1, \dots, b_{300})$ boxes to obtain the distribution density map given by:
\begin{equation}
    \widehat{p}_{n}(b)=\frac{1}{n h}\cdot \sum_{i=1}^{n} K\left(\frac{B_{i}-b}{h}\right)
    \label{eq:kde}
\end{equation}
where $K(b)$ is called the kernel function that is generally a smooth, symmetric function such as a Gaussian and $h>0$ called the smoothing bandwidth controls the amount of smoothing. 

We represent the distribution density map as multiple continuous probability density curves on the image.
The KDE score of a point is computed as the log-likelihood of that point under the KDE model. 
The score reflects the likelihood that any given box has been drawn from the learned probability distribution.
The higher the KDE score is, the more the given box matches the distribution.

\begin{figure}[]
\sidecaption
    \subfloat[\centering Consistent (PCKDE=0.934)]{\includegraphics[width=0.32\hsize]{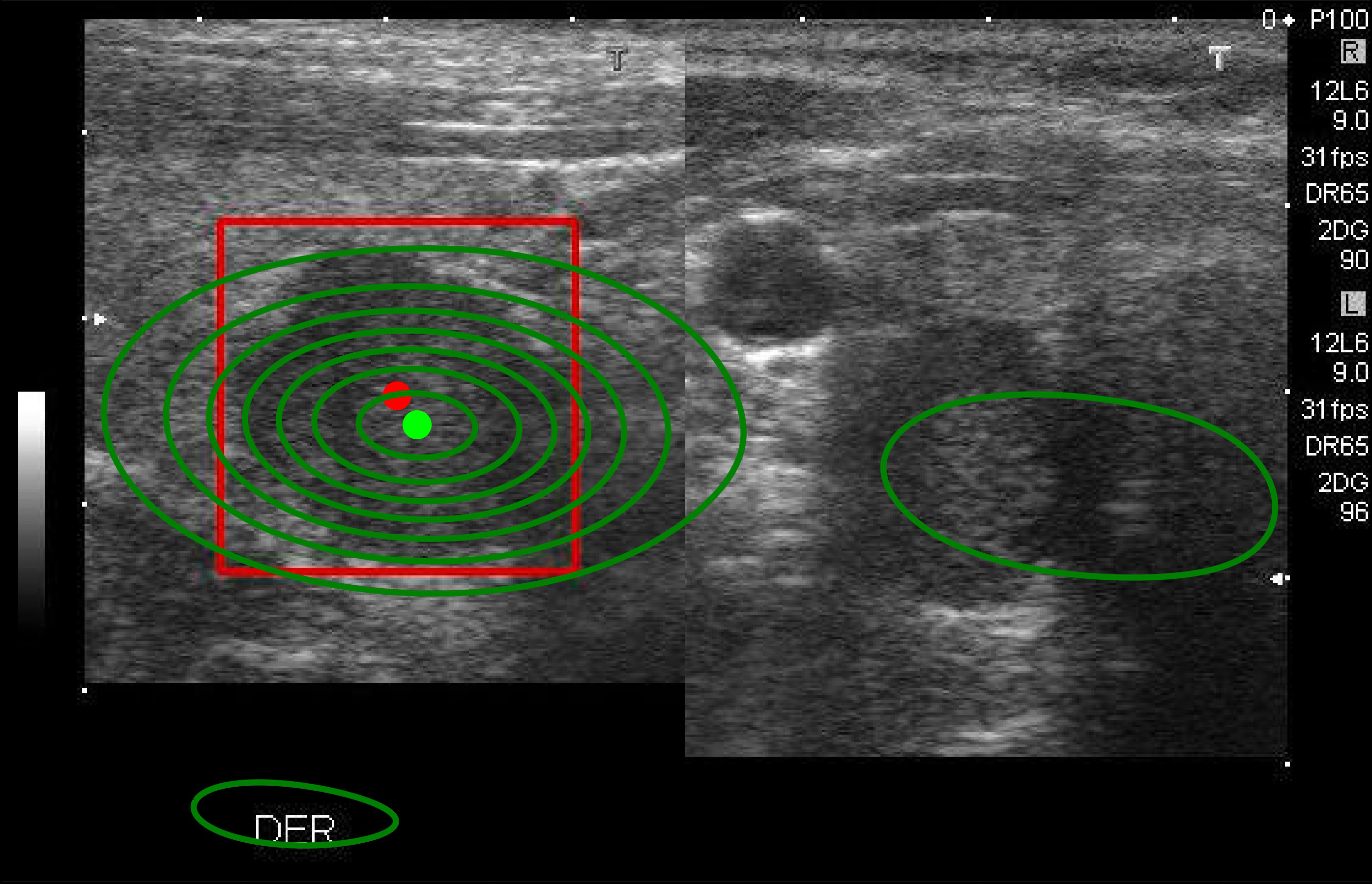}\label{fig:kde_exp}}
    \hfill
    \subfloat[\centering Non-consistent (PCKDE=0.297)]{\includegraphics[width=0.32\hsize]{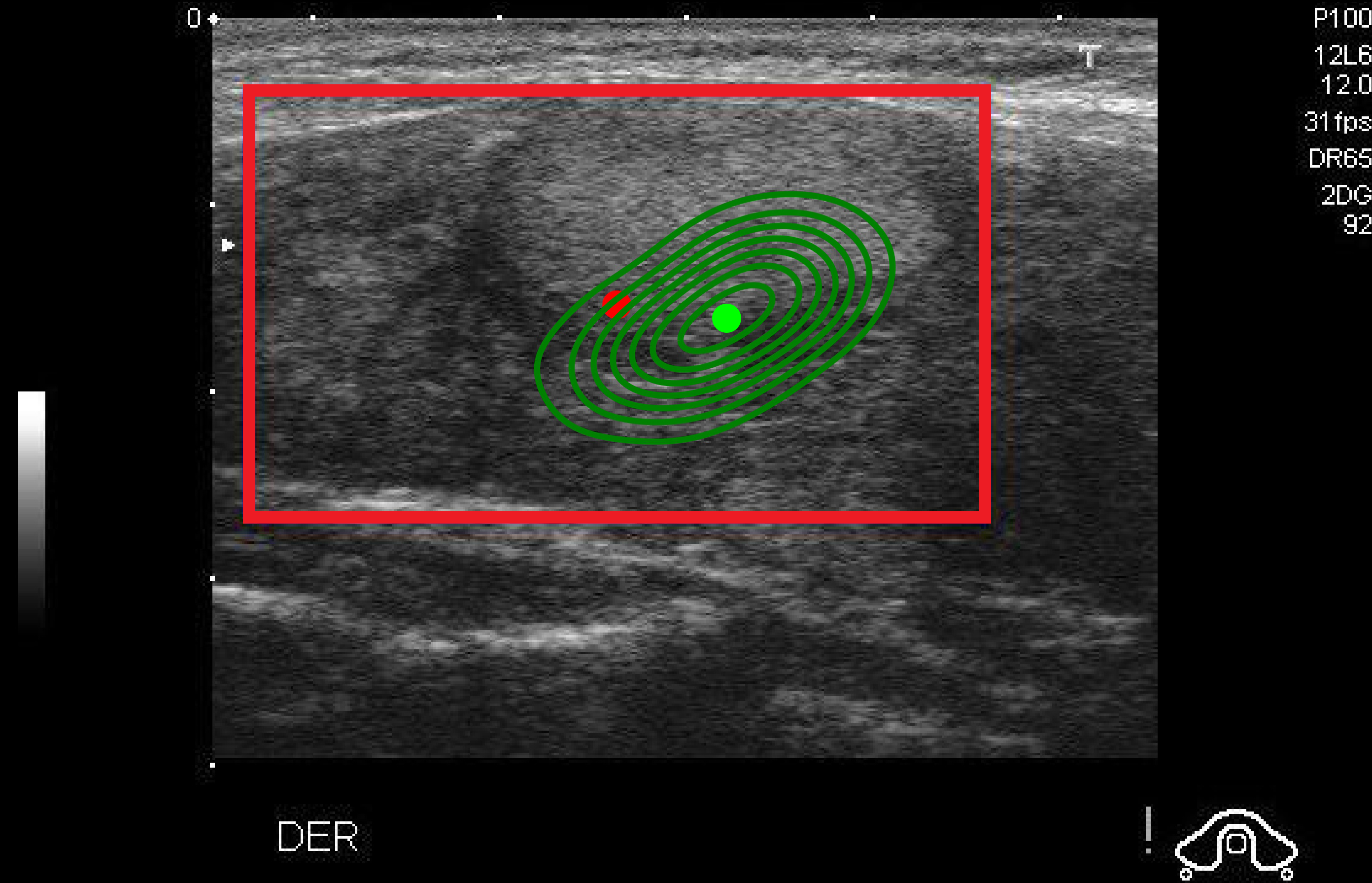}\label{fig:kde_exp_fn}}
    \caption{(a) Consistent: the model detection's center point (red) is mostly at the same location as the point achieving the highest KDE score (green). (b) Non-consistent: the model detection's center point is far from the point having the highest KDE score.}
\label{fig:kde_tp} 
\end{figure}

The prediction's consistency KDE (PCKDE) is computed as the ratio between the KDE score of the final box's center point detected by the model and the KDE score of the point achieving the highest probability value on the distribution density map.
Finally, we set the threshold as 0.5 to grade the PCKDE score. 
If the PCKDE score exceeds 0.5, the model's detection is considered consistent, as shown in Fig.~\ref{fig:kde_exp} and non-consistent in Fig.~\ref{fig:kde_exp_fn}.

\subsubsection{Density map (DM)}\label{subsubsec:density-map}
\begin{figure}[]
\sidecaption
    \subfloat[]{\includegraphics[width=0.32\hsize]{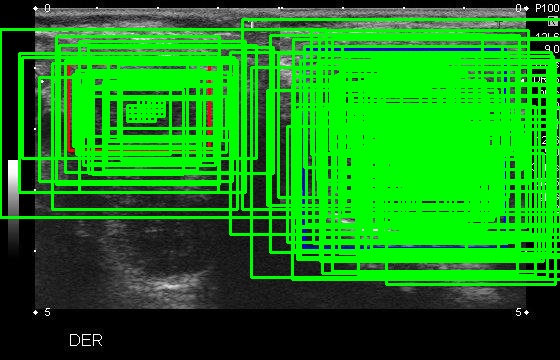}\label{fig:density_map_all_boxes}}
    \hfill
    \subfloat[]{\includegraphics[width=0.32\hsize]{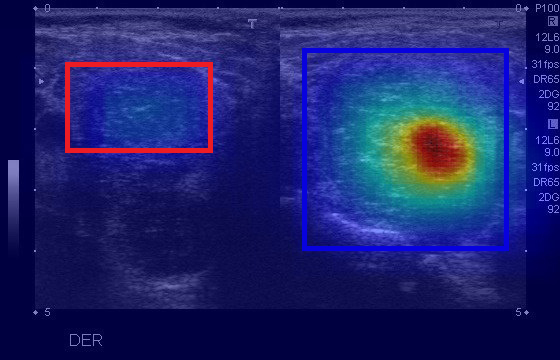}\label{fig:density_map}}
\caption{(a) DM extracts all model's regional proposals after the first stage. (b) DM's explanation as a saliency map.}
\end{figure}

DM is commonly used in crowd counting context to estimate objects' distribution in an image~\cite{li2020density}.
We propose DM as a new statistic-based method in the context of XAI by extracting the frequency value of each pixel from boxes predicted by the model in region proposal generation, as shown in Fig.~\ref{fig:density_map_all_boxes}.
The pixel's frequency value is calculated by the number of boxes containing that pixel.
For an input image $X\in\mathbb{R}^{3\times H \times W}$, the model detected $n$ boxes after the first stage. 
For any box $k$, let $coor_k$ be the set of coordinates $(i,j)$. 
Thus, DM's output $D$ of $X$ is the $H \times W$ matrix defined as $D_{X}=\sum_{i=1}^{n}B_k$, where $B_k$ is computed:
\begin{equation}
\begin{gathered}
    B_k=[a_{ij}]_{H\times W} \\
    a_{ij}=1_{\text{coor}_k}[(i,j)]
\end{gathered}
\end{equation}

The more focused boxes a pixel has, the redder colors are indicated in the DM's explanation.
In Fig.~\ref{fig:density_map}, the model detects two boxes containing nodules where the blue box is correct with the ground-truth label, while the red box is false. 
The DM's saliency map can explain the blue box with redder colors, indicating that the model focuses on this region to detect the nodule. 

\subsubsection{Detector Randomized Input Sampling for Explanation (D-RISE)}
D-RISE is a method for producing saliency maps that explain the regions of an input image that are important for a particular task or model. 
It is specifically designed to produce saliency maps for object detectors, making it the first method. 
It involves generating a set of binary-bit masks for each superpixel in the input image, upsampling the masks using bilinear interpolation, feeding the samples into a model to obtain bounding boxes and scores for each box, and summing up all the masks using the box scores as weights to produce the final saliency map. 
The regions of the input image that significantly impact the model's prediction appear as darker colors on the saliency map.

\section{Results}
\label{sec:results}
\subsection{Qualitative Evaluation}\label{subsec:qualitative-results}

\subsubsection{Analysis of positive cases (with nodules)}
\begin{figure*}[bth!]
    \centering
    \includegraphics[width=\linewidth]{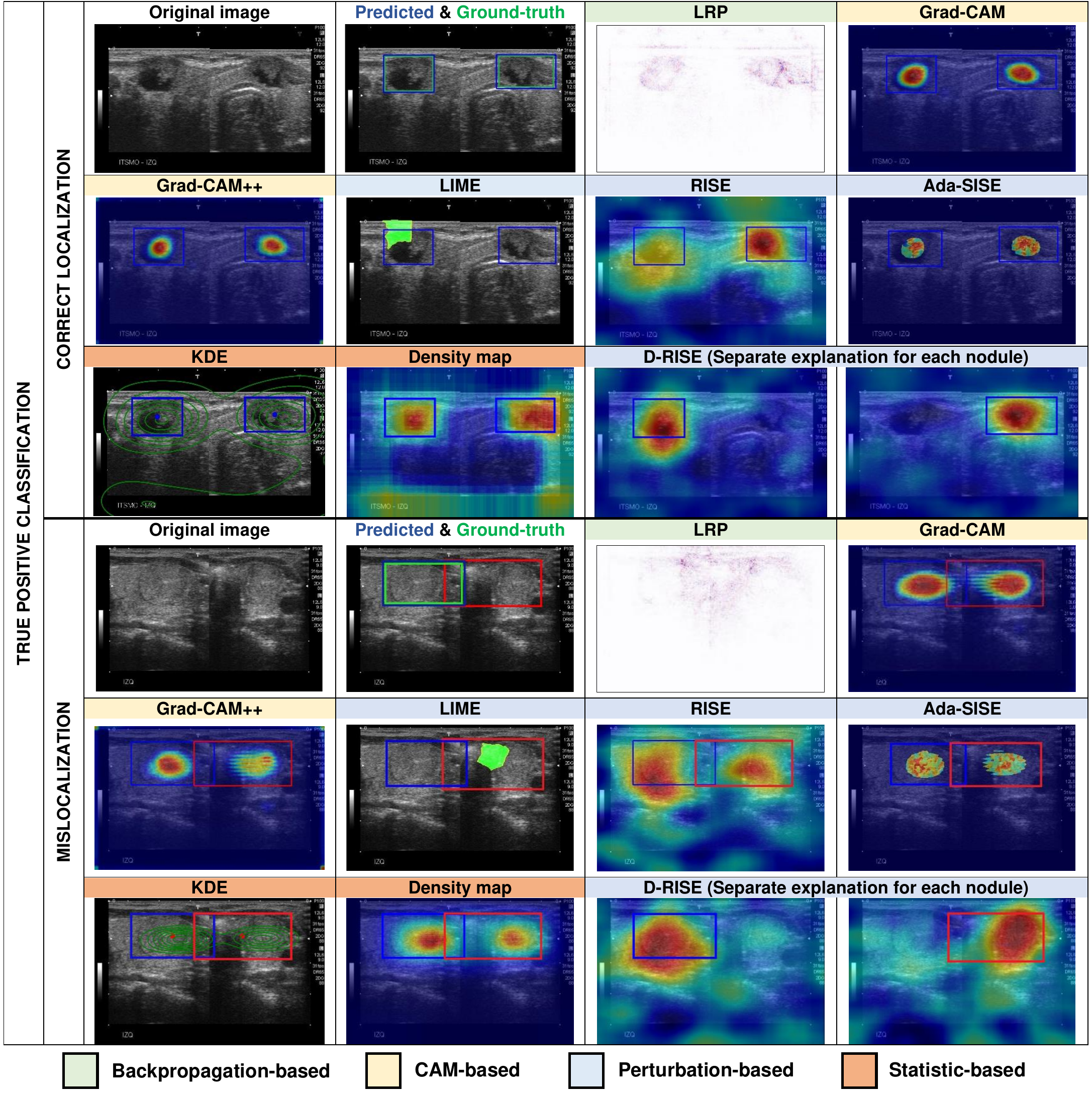}
    \caption{Qualitative comparison between XAI explanations. In the input images, blue boxes are the correct model's predictions, red boxes are the wrong model's predictions, and green boxes are ground truths. The first row is the correct localization case where the predicted box overlaps the ground-truth label. The second row is the mislocalization case where the model predicted two boxes, but the nodule only exists in the left box.}
    \label{fig:qualitative_results}
\end{figure*}

In the positive case, we evaluate XAI explanations on two cases of true positive classification, namely correct localization and mislocalization, as shown in Fig.~\ref{fig:qualitative_results}. 
A correct localization is when the intersection over the union between the detected box for a nodule and the ground-truth box is larger than 0.5. 
All our implemented XAI methods' explanations are applicable in the correct localization case, and their explanations are consistent with the model's detected box in the correct localization case.
Because D-RISE is the only method having access to the localization block, it can give end-users separate explanations for each detected object. 
Meanwhile, all other XAI methods show explanations for all nodules since they explain to end-users why the model classifies an image containing nodules.

To further observe D-RISE's advantages, we consider the mislocalization case where the model predicted two boxes, but the nodule only exists in the left box.
In this case, D-RISE is the only method to show the explanation solely for the correct and incorrect boxes, while others explain both.
Hence, D-RISE's results are more understandable to end-users when each nodule detection needs to be explained.

\subsubsection{Analysis of negative cases (without nodules)}
\begin{figure}[]
\sidecaption
    \includegraphics[width=.6\linewidth]{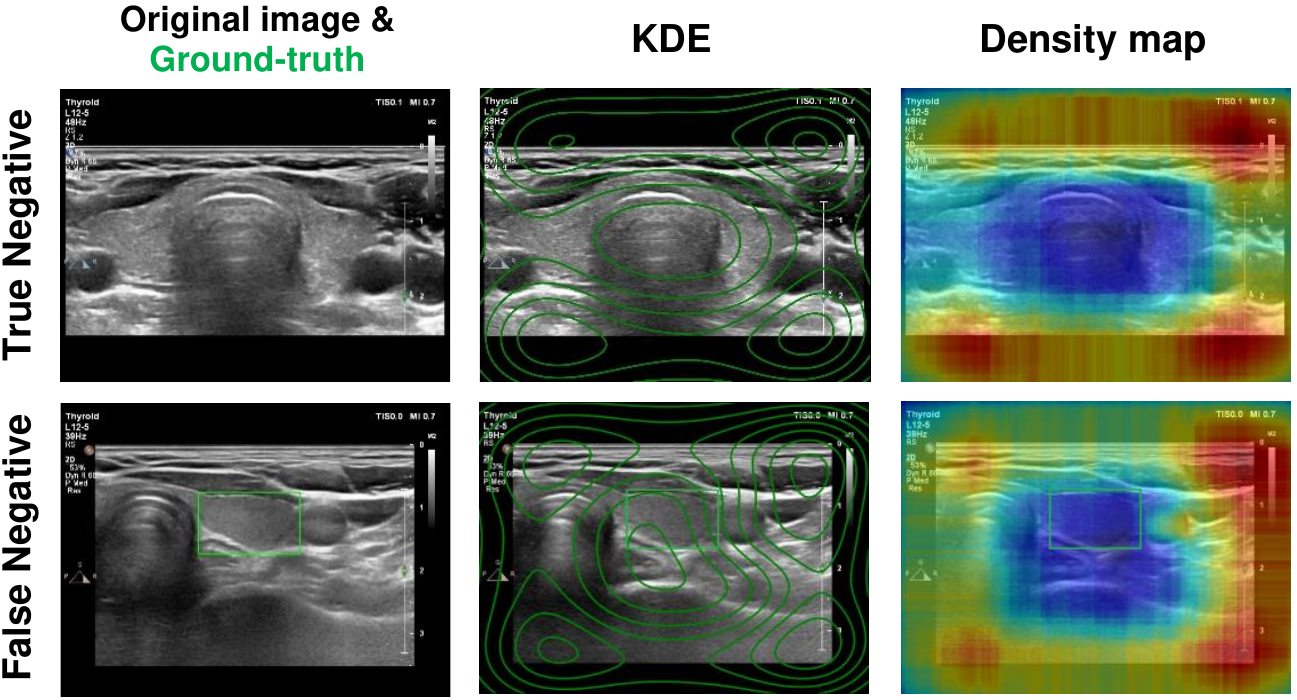}
    \caption{Explanations of KDE and DM for true negative and false negative cases.}
    \label{fig:negative_cases}
\end{figure}

One serious error of AI-assisted thyroid diagnosis is a false negative, where the model fails to detect any nodule.
Hence, our proposed methods, KDE and DM, are the only applicable for explaining negative cases, namely true negative and false negative.
As shown in Fig.~\ref{fig:negative_cases}, KDE estimates the distribution of the model's prediction placed around the image's corner.
The same model's behavior is also reflected in DM, where the saliency map shows hot regions around the image's border.
They both show that the model does not concentrate on any part inside the image, as it does not detect any nodules. 

\subsection{Quantitative Evaluation}\label{subsec:quantitative-results}
Two critical aspects of XAI, plausibility and faithfulness, are evaluated by quantitative metrics.
These metrics are used to justify the model by assessing the extent to which the method satisfies the users by providing superior statistical explanations.
All methods are evaluated on the whole dataset.

\subsubsection{Plausibility Evaluation}
\begin{itemize}
    
\item \textbf{Energy-Based Pointing Game (EBPG)} evaluates XAI methods' precision and denoising ability ~\cite{wang2020score}. 
Extending the traditional Pointing Game, EBPG considers all pixels in the resultant saliency map $S$ for evaluation by measuring the fraction of its energy captured in the corresponding ground truth $G$.

\item \textbf{Intersection over Union (IoU)} analyses the localization ability and meaningfulness of the attributions captured in an explanation map.
Initially, Otsu-based binarization~\cite{otsu1979threshold} is applied to convert ultrasound images into binary images.
We compute the mean IoU between the explanation box and the ground truth box.

\item \textbf{Bounding box (Bbox)} is considered a size-adaptive variant of IoU~\cite{schulz2019restricting}, calculated by selecting the top $N$ pixels in the saliency map significantly influencing the prediction results.
It evaluates the regions captured by the bounding box which contains the object.

\end{itemize}
\subsubsection{Faithfulness Evaluation}
We evaluate XAI's faithfulness with the \textit{Drop} and \textit{Increase}~\cite{fu2020axiom}. 
The original image is perturbed by masking the important areas marked by the explanation.
\begin{itemize}
\item \textbf{Drop} checks the target audience's average predicted drop when using the explanatory as input.
\item \textbf{Increase} measures the number of times the model's confidence increases when using the explanation as input.
\end{itemize}

\begin{table}[]
\centering
\begin{tabular}{cccccccccc}
\toprule
\textbf{Metric} & KDE &  DM & LIME & Grad-CAM & Grad-CAM++ & LRP & RISE & Ada-SISE & D-RISE\\
\midrule
\textbf{EPBG} $\uparrow$ & 29.45 & 21.06 & 10.95 & 48.58 & \textbf{52.11} & 31.17 & 17.04 & \underline{50.31} & 17.52 \\
\midrule
\textbf{BB} $\uparrow$ &49.16 & 40.09 & 14.49 & 60.51   & 47.65 & 38.45 & \underline{62.41} & 55.87 & \textbf{63.42}   \\
\midrule
\textbf{IoU} $\uparrow$ & 18.94 & 18.73 & 10.09 & \underline{45.22}   & \textbf{49.55} & 44.98 & 12.06 & 14.99 & 12.07   \\
\midrule
\textbf{Drop} $\downarrow$ &34.39 & 27.31 & 16.21 & 26.88 & 45.76 & 65.81 & \underline{4.24} & 12.43 & \textbf{2.34} \\
\midrule
\textbf{Inc} $\uparrow$ &32.20 & 27.12 & 27.12 & 18.08 & 9.60 & 4.52& \underline{46.33}& 45.13 & \textbf{53.67}\\
\midrule
\textbf{Time(s)} $\downarrow$ &66 & 28 & 380 & \underline{0.75} & 0.8 & \textbf{0.55} & 245 & 295 & 319 \\
\bottomrule
\end{tabular}

    \caption{Mean accuracy (\%) of quantitative results and computational time of all XAI methods. For each metric, the arrow $\uparrow/\downarrow$ indicates higher or lower scores are better. The best is shown in bold, and the second best is underlined.}
    \label{tab:quantitative_results}
\end{table}

\subsubsection{Result}
Table~\ref{tab:quantitative_results} shows the quantitative comparison of XAI methods concerning their plausibility and faithfulness. 
In detail, CAM-based methods achieved good results with plausibility metrics, especially EPBG and IoU, in a reasonably short time.
While RISE and D-RISE perform better than other methods in terms of faithfulness metrics, such as Drop and Increase, as they faithfully reflect the model's behavior.
Nevertheless, the computational time of D-RISE and RISE in specific and perturbation-based methods, in general, are the highest due to their perturbation process.
LRP achieved the highest computation efficiency, which is its main advantage.

\subsection{Evaluating User Trust}

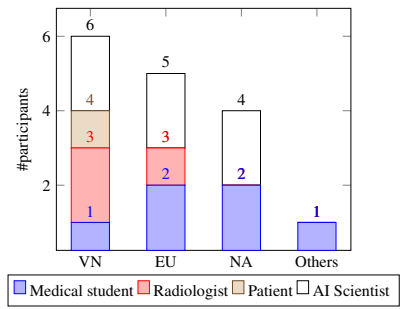
\begin{figure}
\sidecaption
    \resizebox{.45\columnwidth}{!}{%
    \begin{tikzpicture}
    \centering
    \begin{axis}[
        ybar stacked,
    	bar width=20pt,
    	nodes near coords,
        enlargelimits=0.15,
        legend style={at={(0.5,-0.10)},
          anchor=north,legend columns=-1},
        ylabel={\#participants},
        symbolic x coords={VN, EU, NA, Others},
        xtick=data,
        ]
    \addplot+[ybar] plot coordinates {(VN,1) (EU,2) 
      (NA,2) (Others,1)};
    \addplot+[ybar] plot coordinates {(VN,2) (EU,1)
      (NA, 0) (Others,0)};
    \addplot+[ybar] plot coordinates {(VN,1) (EU,0) 
      (NA,0) (Others,0)};
    \addplot+[ybar, fill=white] plot coordinates {(VN,2) (EU,2) (NA,2) (Others,0)};
    \legend{\strut Medical student, \strut Radiologist, \strut Patient, \strut AI Scientist}
    \end{axis}
    \end{tikzpicture}
    }
    \caption{The number of participants from Vietnam (VN), European countries (EU), North American countries (NA), and other countries. Participants are medical students, radiologists, patients, and AI scientists.}
    \label{fig:participants}
\end{figure}

An essential question of XAI applications is whether explanations can build end-users' trust in the AI system.  
We invited 16 participants with previous experience and familiarity with ultrasound images and AI models, namely three radiologists, one patient, six medical students, and 6 AI scientists from different countries, to take a survey.
Six participants were self-reported as from Vietnam, five were from European countries (Germany, Romania, Ukraine), four were from North American countries (United States, Canada), and one was from Pakistan, as shown in Fig.~\ref{fig:participants}.
In this survey, the trust of humans towards XAI's explanation is evaluated in terms of understandability, where this explanation helped us understand why the model behaved as it did.

\begin{figure}
    \subfloat[]{\includegraphics[width=0.49\hsize]{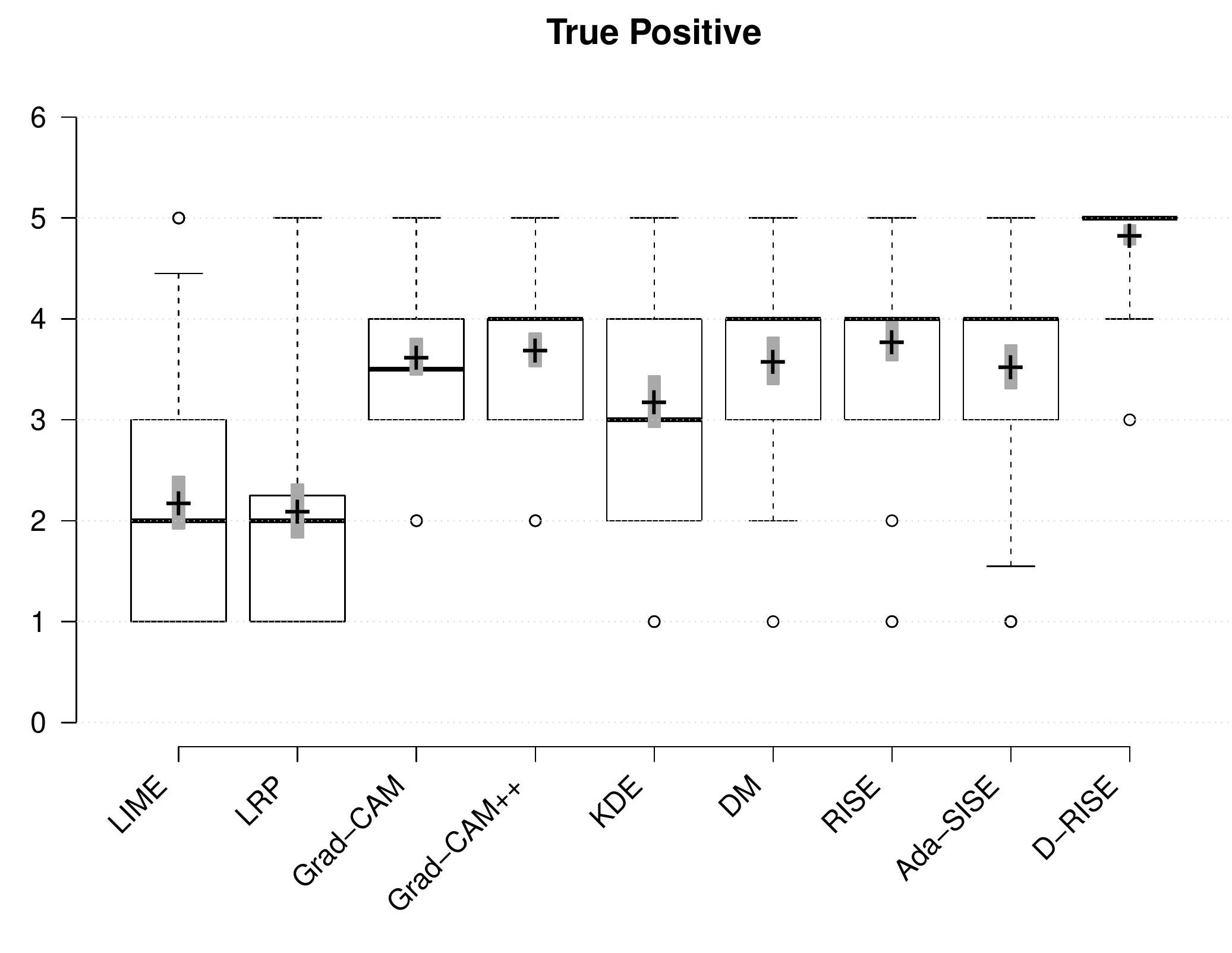}\label{fig:tp_box}}
    \hfill
    \subfloat[]{\includegraphics[width=0.49\hsize]{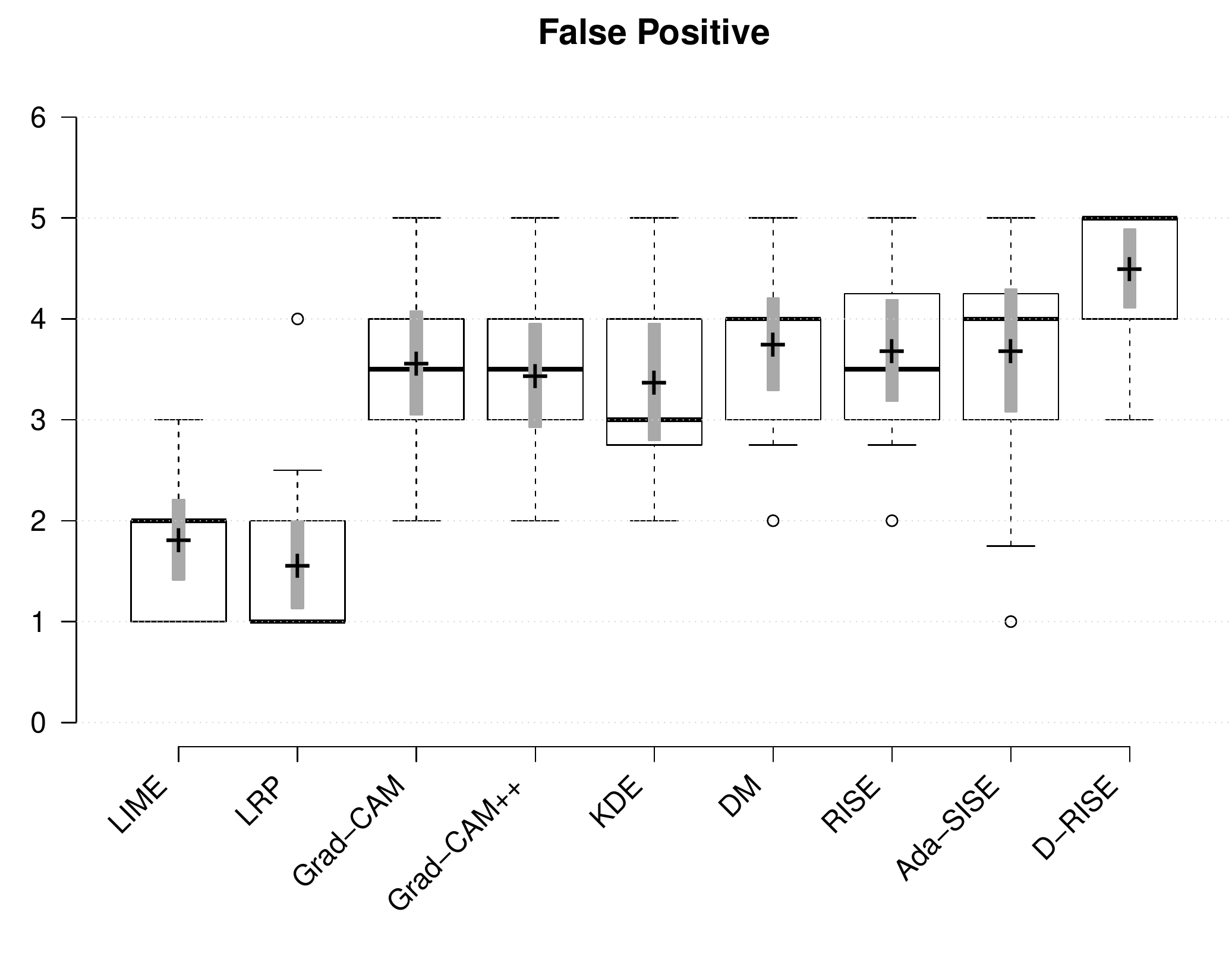}\label{fig:fp_box}}
    \\
    \sidecaption
    \subfloat[]{\includegraphics[height=0.33\hsize]{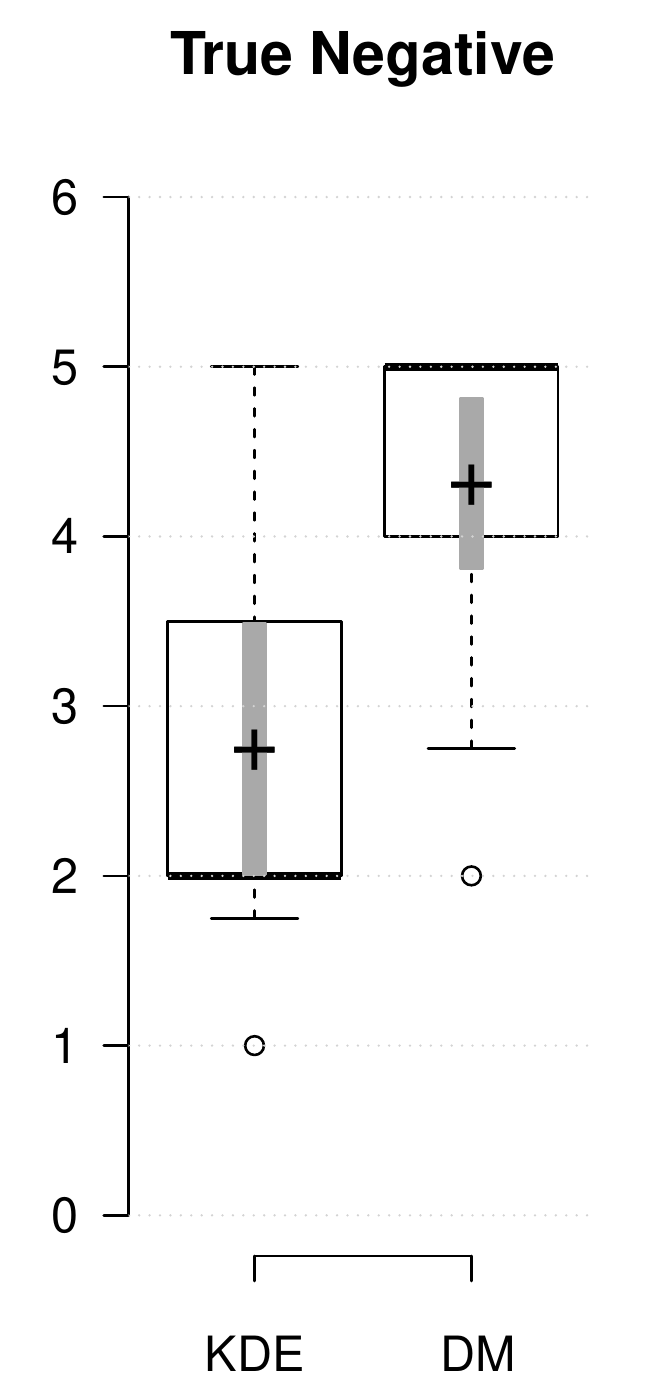}\label{fig:tn_box}}
    \hspace{6mm}
    \subfloat[]{\includegraphics[height=0.33\hsize]{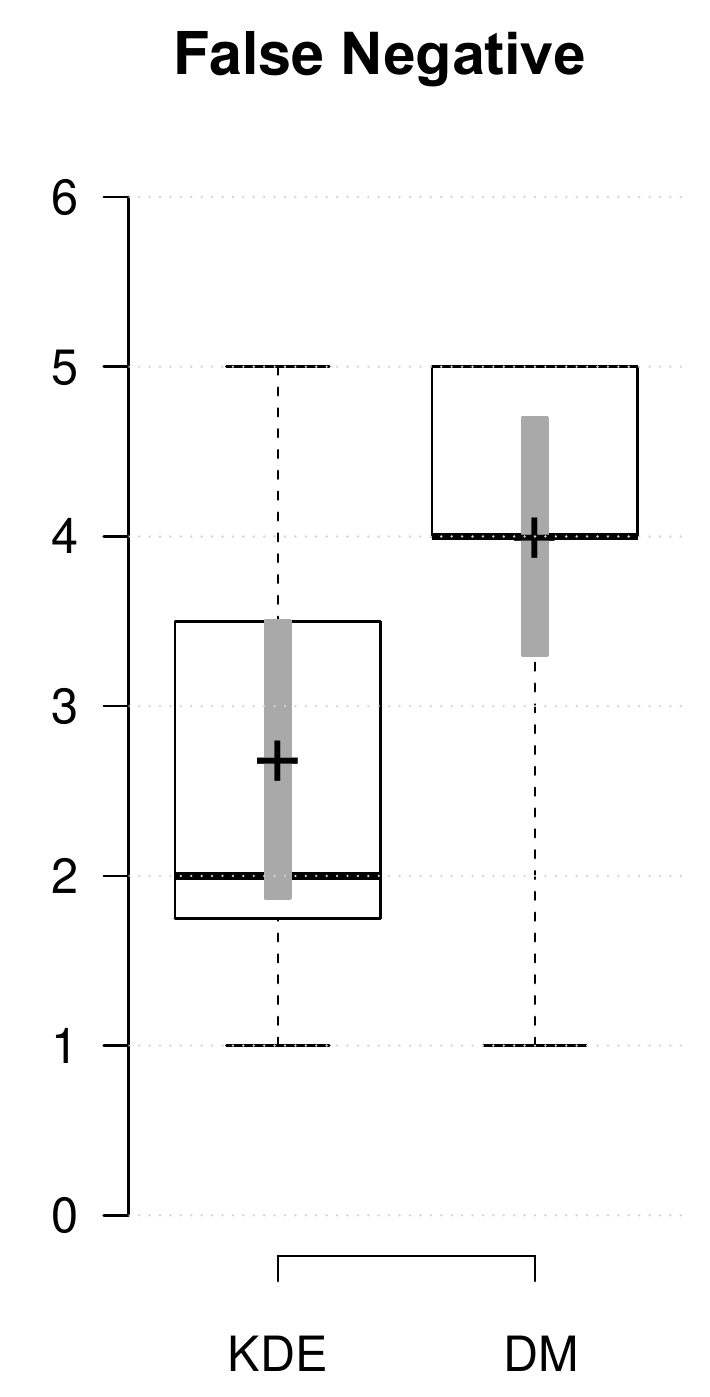}\label{fig:tn_box}}
    \caption{Participants ratings of XAI methods in (a) True Positive, (b) False Positive, (c) True Negative, and (d) False Negative.}
    \label{fig:user_trust} 
\end{figure}

All participants answered seven questions categorized into four prediction cases, namely True Positive (TP), False Positive (FP), True Negative (TN), and False Negative (FN).
In detail, TP cases have four questions for the following specific cases: one question for one nodule correctly detected, one question for two nodules correctly detected (correct localization), and two questions for one nodule correctly detected but another falsely detected (mislocalization).
While there is one question for each FP, TN, and FN. 
Each question represents a specific case of the model's classification and localization where the model's prediction, ground truth, and explanations of applicable XAI methods are shown.
Participants rated their trust of each XAI's explanation on a labeled, 5-point Likert scale, ranging from 1 (very unlikely) to 5 (very likely). 
The box plots in Fig.~\ref{fig:user_trust} statistically report the score of participants' ratings of XAI methods on four prediction cases. 
All box plots use the Altman whiskers to display the spread of participants' ratings, which can be particularly useful for showing outliers.

In general, XAI methods using a saliency map as the explanation have consistently higher understandability with higher medians towards humans than others in all cases.
D-RISE gains the most trust from users in TP and FP cases, where its interquartile ranges are from 4 to 5.
While in TN and FN cases, DM overall surpasses KDE.
Still, despite not having a user-friendly explanation, KDE's maximum whisker reaches 5, which means that it still gains some high trust from users.
Also, note that all methods contain low-rating outliers, which means that probably few users are still confused with explanations.
Thus, the future of different explanation types is still wide-open.

\section{Conclusion}
\label{sec:conclusion}
We applied several XAI methods and proposed two new statistic-based methods, namely KDE and Density map, in the context of XAI to explain the model's predictions on the Vietnamese thyroid ultrasound dataset.
Our implemented and proposed XAI methods can cover all prediction cases with high consistency with the object detector and doctors' knowledge.
Consequently, according to our evaluations and surveys, we recommend end-users use Grad-CAM++ as the default method since it requires a very short time to explain plausibly per case. At the same time, D-RISE is suitable when we require explicitly separate explanations for each nodule due to its faithfulness but high computational time.
In future works, we would like to integrate XAI methods into the diagnosis process in real-time scenarios, so the transparency of AI decisions to doctors and patients can be improved.
Also, we aim to conduct a more comprehensive survey that spans multiple countries and continents involving more various user subjects to increase the representativeness and reliability of XAI in the medical field.

\end{document}